\documentclass[letterpaper]{article} 
\pdfoutput=1
\usepackage{aaai25}  
\usepackage{times}  
\usepackage{helvet}  
\usepackage{courier}  
\usepackage{url}  
\usepackage[pdftex]{graphicx} 
\usepackage{algorithm}
\usepackage{algpseudocode}
\usepackage{natbib}  
\usepackage{caption} 
\usepackage{newfloat}
\usepackage{listings}
\usepackage{xcolor}
\DeclareCaptionStyle{ruled}{labelfont=normalfont,labelsep=colon,strut=off} 
\floatstyle{ruled}
\newfloat{listing}{tb}{lst}{}
\floatname{listing}{Listing}
\usepackage{enumitem}
\usepackage{amsmath, amssymb, amsthm}
\usepackage{booktabs}
\usepackage{multirow}

\title{Graph Coarsening via Supervised Granular-Ball for Scalable Graph Neural Network Training}
\author {
    Shuyin Xia\textsuperscript{\rm 1,\rm 2,\rm 3},
    Xinjun Ma\textsuperscript{\rm 2,\rm 3},
    Zhiyuan Liu\textsuperscript{\rm 2,\rm 3},
    Cheng Liu\textsuperscript{\rm 2,\rm 3},
    Sen Zhao\textsuperscript{\rm 1,\rm 2,\rm 3}\thanks{Corresponding author.},
    Guoyin Wang\textsuperscript{\rm 4},
}
\affiliations{
    \textsuperscript{\rm 1}Key Laboratory of Big Data Intelligent Computing\\

    \textsuperscript{\rm 2}Chongqing Key Laboratory of Computational Intelligence\\
    \textsuperscript{\rm 3}Chongqing University of Posts and Telecommunications, Chongqing, China\\
    \textsuperscript{\rm 4}National Center for Applied Mathematics in Chongqing, Chongqing Normal University, Chongqing 401331, China\\
    \{xiasy,zhaoseng\}@cqupt.edu.cn;
    \{s220201072,s230231079,s230201067\}@stu.cqupt.edu.cn
}

\usepackage{bibentry}

\begin{document}

\maketitle

\begin{abstract}
Graph Neural Networks (GNNs) have demonstrated significant achievements in processing graph data, yet scalability remains a substantial challenge. To address this, numerous graph coarsening methods have been developed. However, most existing coarsening methods are training-dependent, leading to lower efficiency, and they all require a predefined coarsening rate, lacking an adaptive approach. In this paper, we employ granular-ball computing to effectively compress graph data. We construct a coarsened graph network by iteratively splitting the graph into granular-balls based on a purity threshold and using these granular-balls as super vertices. This granulation process significantly reduces the size of the original graph, thereby greatly enhancing the training efficiency and scalability of GNNs. Additionally, our algorithm can adaptively perform splitting without requiring a predefined coarsening rate. Experimental results demonstrate that our method achieves accuracy comparable to training on the original graph. Noise injection experiments further indicate that our method exhibits robust performance. Moreover, our approach can reduce the graph size by up to 20 times without compromising test accuracy, substantially enhancing the scalability of GNNs. 
\begin{links}
     \link{Code}{https://github.com/mxjun17/Granular-Ball-Node-Classifction}
\end{links}

\end{abstract}

\section{Introduction}

In recent years, graph neural networks (GNNs) have shown significant potential across various domains, including social network analysis, drug discovery, finance, and material science \cite{20,3,4,5,32,33,35}. The increasing prevalence of large-scale graph data offers richer information and larger training sets for learning algorithms. However, this also amplifies the challenge of processing such data, demanding substantial computational resources and escalating costs, particularly during model training and parameter tuning \cite{21}. A practical solution is to simplify or reduce the graph, which not only accelerates GNNs but also enhances graph data analysis tasks like storage, visualization, and retrieval.

\begin{figure}[tb!]
  \centering
  \includegraphics[width=0.75\linewidth]{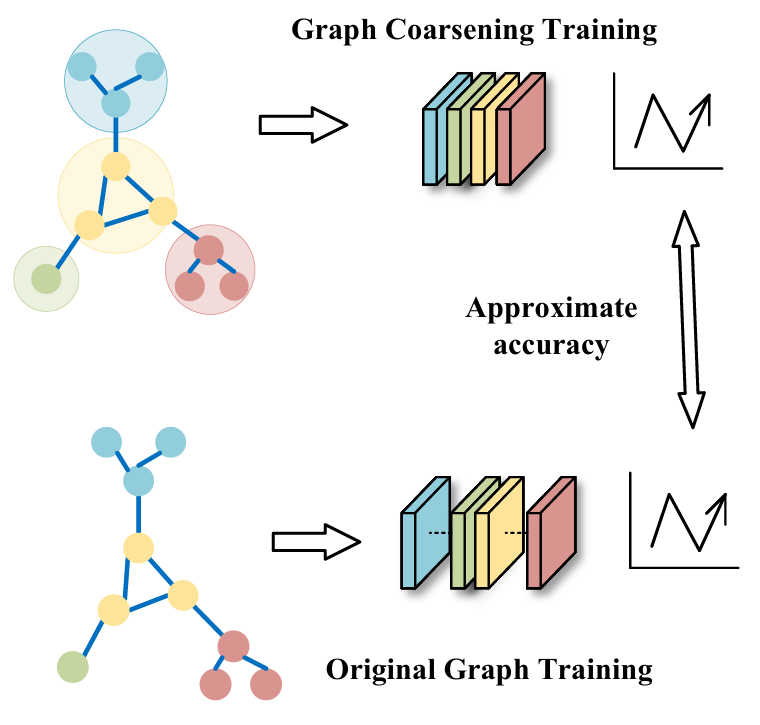}
  \caption{Comparing original graph training with coarsening training. The coarsening approach reduces graph complexity while preserving key structural and label information, improving graph neural network training efficiency.}
  \label{fig:coarsening}
\end{figure}

Training graph neural networks (GNNs) often incurs significant computational costs. To mitigate this, dataset condensation or distillation has been explored, constructing smaller synthetic datasets for training \cite{24,25}. However, these methods may demand high network performance and require precise compression control to be effective. Graph coarsening offers a solution by reducing large graphs into smaller structures while preserving key attributes, based on algebraic multigrid theory. Widely used in graph partitioning, summarization, and machine learning \cite{6,17,18,7,8}, coarsening methods often focus on preserving spectral features \cite{31,27,28,23,22}. Despite their benefits, the high memory and computational costs of spectral methods can limit their scalability. While applied in deep learning \cite{34,19,21,29,26,49,61}, coarsening faces challenges like excessive information loss, which can degrade model performance.

Figure 1 illustrates the graph coarsening training process, highlighting the critical importance of effective coarsening, as it inevitably leads to the loss of some original graph information. While training-dependent graph coarsening methods like CMGC\cite{61} and GCOND\cite{24} achieve high accuracy, they suffer from inefficiency and prolonged training times. In contrast, preprocessing-based methods like SCAL\cite{21} offer better efficiency but lack the guidance of label information, resulting in slightly lower accuracy. Moreover, most existing graph coarsening techniques are not adaptive and require a predefined coarsening ratio.

To address these challenges, we propose a Supervised Granular-Ball Graph Coarsening \textbf{(SGBGC)} method, which optimizes information preservation by integrating node label information with graph structural characteristics and introduces an adaptive coarsening mechanism. This adaptability enhances the flexibility of graph coarsening. Our approach employs an innovative granular-ball generation mechanism, initially partitioning granular-balls based on node label similarity, followed by granularity adjustments to refine these granular-balls into high-quality super nodes.

We conducted extensive experiments on real datasets to demonstrate the effectiveness of the SGBGC method. Our approach outperforms all existing graph coarsening architectures for node classification and achieves state-of-the-art results on five benchmark datasets for node classification. Detailed analyses and robustness studies further illustrate the superiority of our methods. In summary, the main contributions are:

\begin{itemize}[leftmargin=*]
    \item We propose a novel Supervised Granular-Ball Graph Coarsening (SGBGC) method that adaptively utilizes granular-balls of varying sizes to represent the sample space, maintaining crucial graph structural properties while enhancing the scalability of Graph Neural Network (GNN) models.
    
    \item SGBGC reduces time complexity by avoiding the need to calculate distances between all node pairs within the graph. Additionally, Our method has demonstrated superior noise resistance in robustness experiments, outperforming existing graph coarsening techniques.
    
    \item Extensive testing and comparisons on real-world datasets have confirmed the effectiveness and accuracy of SGBGC. Our method has achieved higher precision in experiments, validating its efficacy for GNN training and inference.
\end{itemize}

\begin{figure*}[tb!]
  \centering
  \includegraphics[width=\linewidth]{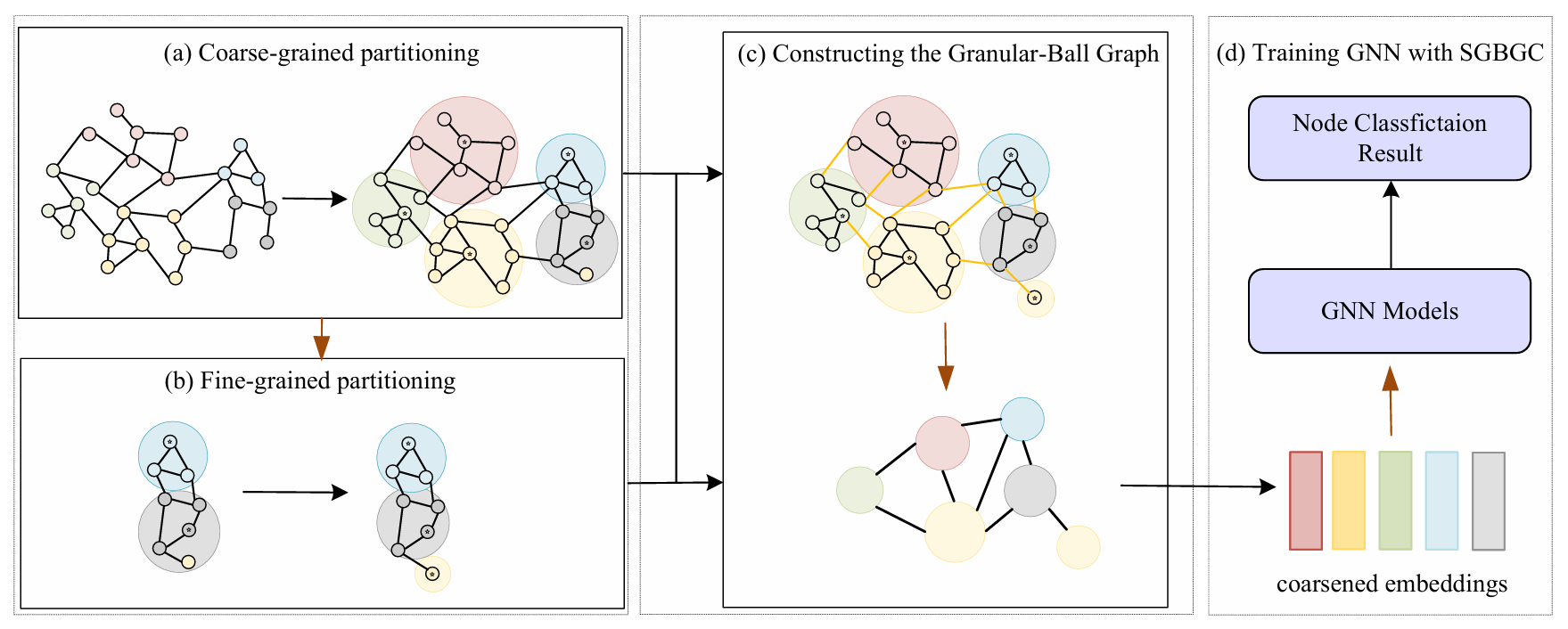}
  \caption{The overview of the SGBGC architecture. Our method consists of four stages: coarse partitioning, fine-grained splitting, granular-ball graph construction, and GNN training on the coarsened graph. In the coarse partitioning stage, the graph is divided into $\sqrt{N}$ clusters, where $N$ is the total number of nodes, with each cluster evenly divided based on label categories. The fine-grained splitting stage divides clusters further based on granular-ball quality until the specified threshold is reached. Finally, the granular-ball graph is constructed and used for GNN training to produce coarsened embeddings.}
  \label{fig:SGBGC_architecture}
\end{figure*}

\section{Related Works}
\label{appendix:a}

\subsection{Graph Coarsening and Condensation}
During the training of Graph Neural Networks (GNNs), due to the typically high computational costs, methods such as graph coarsening and graph condensation are increasingly gaining attention. These methods aim to create smaller synthetic datasets for training GNNs, thereby reducing costs to some extent. Among these, SCAL \cite{21} demonstrated scalable training of GNNs through graph coarsening techniques, proving that even with a reduction of graph nodes to one-tenth of the original size, there was no significant impact on classification accuracy. FGC \cite{26} designed an innovative optimization-based graph coarsening framework, which takes the adjacency matrix and node features as inputs, and jointly learns the coarsened versions of the graph and feature matrices to ensure the desired properties while optimizing graph representation. \cite{24,25} proposed the GCOND graph compression method, which not only reduces the number of nodes but also compresses node features and structural information in a supervised manner. CMGC \cite{61} focuses on preserving graph convolution operations while enhancing existing coarsening techniques by merging nodes with similar structures. It is also a training-dependent coarsening method, delivering state-of-the-art performance.

\subsection{Granular-Ball Computing}
\label{appendix:a.3}
Building on the theoretical foundations of traditional granularity computation and integrating the human cognitive mechanism of 'macro-first' \cite{2}, Wang \cite{32} pioneered the concept of multi-granularity cognitive computation. Following this approach, Xia \cite{36} proposed an effcient, robust and interpretable computational method known as granular-ball computing. Compared to traditional methods that input data at the finest granularity of a point, granular-ball computing uses granular-balls to cover and represent data and takes granular-balls as inputs, offering high effciency, robustness. A granular-ball is defined by a point set covered by the ball, having a centroid and a radius in the standard formation, such as in granular-ball classification. Since its introduction, although the theory of granular-ball computing has not been developed for long, it has been widely applied across multiple domains of artificial intelligence, giving rise to various granular-ball generation techniques and related computational models. Specific examples include granular-ball classifiers \cite{45,国外粒球svm,分类2}, granular-ball clustering \cite{53,pami,63,ICDE2}, granular-ball sampling methods \cite{41}, granular-ball rough set\cite{42,40,zhang2023incremental}, granular-ball three way decision\cite{gbroughsets1,gbroughsets2}, and developments like granular-ball reinforcement learning \cite{62}. Furthermore, several applications have demonstrated the effectiveness of granular-ball representation, such as granular-ball representation in text adversarial defense\cite{wang2024text}, label noise combating\cite{wang2024gbrain,dai2024granular}.

\section{Preliminaries}

\subsection{Graph Coarsening}
We represent a graph with \(N\) nodes and \(M\) edges as \(\mathcal{G} = (\mathcal{V}, \mathcal{E}, \mathbf{X})\), where \(\mathcal{V}\) is the set of nodes, \(\mathcal{E}\) is the set of edges, \(\mathbf{X}\) represents the features of the nodes, and \(\mathbf{Y}\) represents the set of node labels. All graph data in this paper are undirected graphs. We use \(\mathbf{A} \in \{0, 1\}^{N \times N}\) to denote the adjacency matrix of \(\mathcal{G}\), i.e., the \((i, j)\)-th entry in \(\mathbf{A}\) is 1 if and only if there is an edge between \(v_i\) and \(v_j\).
Given an undirected graph \(\mathcal{G}\), graph coarsening seeks to find a coarser graph \(\mathcal{G}' = (\mathcal{V}', \mathcal{E}', \mathbf{X}')\) that approximates \(\mathcal{G}\), where \(\mathcal{V}'\) is the set of super nodes obtained after coarsening, \(\mathcal{E}'\) is the set of super edges, \(\mathbf{X}'\) represents the features of the super nodes, and \(\mathbf{A}'\) is the adjacency matrix of the coarser graph. The node count in \(\mathcal{G}'\) is \(n < N\) and the edge count is \(m < M\). By computing a partition \(\mathcal{P} = \{C_1, C_2, \ldots, C_n\}\), we can construct \(\mathcal{G}'\), where each cluster \(C_i\) corresponds to a super node in \(\mathcal{G}'\), and the connections between these super nodes \(C_i\) and \(C_j\) form the super edges, i.e., a tightly connected subset of nodes from the original graph.

\subsection{Traditional Granular-Ball Computing}
The core concept of granular-ball computing involves partitioning the sample space into manageable segments using granular-balls. These granular-balls serve as the basic units for learning and facilitate multi-granularity representation and computation. The traditional granular-ball is defined by a center point and a radius, encompassing all points within the specified radius. Formally, a granular-ball centered at a point \( \mathbf{c} \in \mathbb{R}^d \) with radius \( r \) is defined as:
\begin{equation}
\mathcal{GB} = \{ \mathbf{x} \in \mathbb{R}^d : \| \mathbf{x} - \mathbf{c} \| \leq r \},
\end{equation}
where \( \| \mathbf{x} - \mathbf{c} \| \) denotes the Euclidean distance between any point \( \mathbf{x} \in \mathbb{R}^d \) and the center \( \mathbf{c} \in \mathbb{R}^d \). The radius \( r \) is determined based on the distribution of data points within the cluster, providing a comprehensive representation of the data within that region. To ensure that the entire dataset \( \mathcal{D} \subseteq \mathbb{R}^d \) is covered, the dataset is partitioned into \( n \) granular-balls. The collection of these granular-balls, denoted as \( \mathcal{GB}_{s} = \{\mathcal{GB}_1, \mathcal{GB}_2, \ldots, \mathcal{GB}_n\} \), is used such that \( \mathcal{D} \subseteq \bigcup_{i=1}^n \mathcal{GB}_i \). This ensures that every data point in \( \mathcal{D} \) is included in at least one granular-ball, facilitating complete coverage of the sample space.
By using granular-balls instead of individual data points, granular computing simplifies data representation and reduces computational complexity. This structure forms a cluster that can be effectively used to represent spatial relationships in data.

\section{Methodology}

\subsection{Overview}
Figure 2 presents an overview of our proposed SGBGC framework.
In contrast to traditional granular-ball computing, which primarily focuses on abstract data granulation, in graph-based scenarios, a granular-ball \( \mathcal{GB} \) integrates both structural and feature-specific properties of graph data.

\paragraph{\textbf{Definition 1: Granular-Ball on Graph}}\mbox{}\\
Consider an undirected graph \( \mathcal{G} = (\mathcal{V}, \mathcal{E}, \mathbf{X}, \mathbf{A}) \), where \( \mathcal{V} \) is the set of nodes with \( |\mathcal{V}| = N \) representing the total number of nodes, \( \mathcal{E} \) is the set of edges with \( |\mathcal{E}| = M \) indicating the total number of edges, \( \mathbf{X} \) is the matrix of node features, and \( \mathbf{A} \) is the adjacency matrix of the graph.
A granular-ball \( \mathcal{GB} \) is defined as a connected subgraph of \( \mathcal{G} \), represented by:
\begin{equation}
\mathcal{GB} = (\mathcal{V}_{gb}, \mathcal{E}_{gb}, \mathbf{X}_{gb}, \mathbf{A}_{gb})
\end{equation}

Here, \( \mathcal{V}_{gb} \subseteq \mathcal{V} \) represents the subset of nodes in the granular-ball, \( \mathcal{E}_{gb} \subseteq \mathcal{E} \) represents the subset of edges among nodes within \( \mathcal{V}_{gb} \), \( \mathbf{X}_{gb} \subseteq \mathbf{X} \) represents the subset of features corresponding to \( \mathcal{V}_{gb} \), and \( \mathbf{A}_{gb} \) represents the adjacency matrix corresponding to \( \mathcal{V}_{gb} \). The granular-ball \( \mathcal{GB} \) must be a connected subgraph, which means for any two nodes \( v_i, v_j \in \mathcal{V}_{gb} \), there does not exist a situation where \( d(v_i, v_j) = \infty \). Here, \( d(v_i, v_j) \) represents the shortest path distance (in terms of the number of edges) between nodes \( v_i \) and \( v_j \) in the graph.
The center node \( c \) of a granular-ball is defined as the node within \( \mathcal{V}_{gb} \) with the maximum degree:
\begin{equation}
c = \mathop{\arg\max}_{v \in \mathcal{V}_{gb}} \deg_{gb}(v),
\end{equation}
where \( \deg_{gb}(v) \) is the degree of node \( v \) within the granular-ball \( \mathcal{GB} \), defined as the number of edges connected to \( v \) within \( \mathcal{E}_{gb} \).

\paragraph{\textbf{Definition 2:The  Purity of the Granular-Ball}}\mbox{}\\
The purity \( T \) of a granular-ball \( \mathcal{GB} \) is defined as the ratio of the count of the most frequent label to the total number of nodes within the ball. This ratio quantifies the uniformity of labeling within \( \mathcal{GB} \). Mathematically, it is given by:
\begin{equation}
T = \frac{\max(n_i)}{n}, \quad i = 1, 2, \ldots, k,
\end{equation}
where \( n_i \) is the number of nodes with label \( i \) in \( \mathcal{GB} \), \( n \) is the total number of nodes in \( \mathcal{GB} \), and \( k \) is the number of distinct labels in the dataset. Here, \( T \) falls strictly within the range \( \frac{1}{k} < T \leq 1 \), where \( T = 1 \) indicates that all nodes in \( \mathcal{GB} \) share the same label, representing maximum purity. The lower bound \( \frac{1}{k} \) occurs when the node labels are distributed as evenly as possible across the \( k \) labels, representing the least purity achievable under the constraint that one label must be most frequent.

\subsection{Problem Definition}

In our framework, the original graph \( \mathcal{G} = (\mathcal{V}, \mathcal{E}, \mathbf{X}) \) is transformed into a granular-ball graph, denoted as \( \tilde{\mathcal{G}} = (\tilde{\mathcal{V}}, \tilde{\mathcal{E}}, \tilde{\mathbf{X}}) \), by coarsening the graph into \( n \) granular-balls. Each granular-ball \( \mathcal{GB}_j \) corresponds to a super node in \( \tilde{\mathcal{G}} \), forming the set \( \tilde{\mathcal{V}} = \{\mathcal{GB}_1, \mathcal{GB}_2, \ldots, \mathcal{GB}_n\} \) where \( n \) is the total number of super nodes. The goal is to obtain a coarsened graph \( \tilde{\mathcal{G}} \) that reduces the complexity of the original graph while retaining essential structural and feature information.
In this coarsened graph \( \tilde{\mathcal{G}} \), the adjacency matrix \( \tilde{\mathbf{A}} \) is defined such that an element \( \tilde{\mathbf{A}}_{ij} \) equals 1 if there is an edge between super nodes \( \mathcal{GB}_i \) and \( \mathcal{GB}_j \); otherwise, \( \tilde{\mathbf{A}}_{ij} = 0 \). Each node \( v_i \) in the original graph has a label \( y_i \), and in the node classification task, the classifier is initially aware of the labels of a subset of nodes \( \mathcal{V}_L \). The goal of node classification is to infer the labels of nodes in \( \mathcal{V} \setminus \mathcal{V}_L \) by learning a classification function from the coarsened graph \( \tilde{\mathcal{G}} \), effectively utilizing the reduced graph structure to predict labels with reduced computational resources while maintaining or even enhancing prediction accuracy.

\subsection{Supervised Granular-Ball Graph Coarsening}

The SGBGC method comprises two main stages: initial coarse partitioning and fine-grained binary splitting. In the initial stage, the goal is to simplify the graph structure by establishing a uniform distribution of granular-balls based on node labels and their degree of connectivity. Starting with the highest degree nodes as centers, \(\alpha = \sqrt{N}\) centers are chosen, where \(N\) is the total number of nodes. The value of \(\sqrt{N}\) is an empirical choice, referenced from previous work \cite{50,60}, which has been shown to balance the trade-off between computational complexity and accuracy. These \(\alpha\) centers are then evenly distributed among the \(k\) label categories, resulting in each category initially having \(\alpha/k\) centers. 

This setup forms the preliminary granular-ball distribution,
\begin{equation}
\mathcal{GB}_{\text{init}} = \{\mathcal{GB}_1, \mathcal{GB}_2, \ldots, \mathcal{GB}_{\alpha}\},
\end{equation}
which covers the entire graph dataset. The choice of high-degree nodes ensures that the initial centers are well-connected, facilitating effective coverage of the graph. Each \(\mathcal{GB}_i\) (\(1 \leq i \leq \alpha\)) in \(\mathcal{GB}_{\text{init}}\) undergoes binary splitting in the fine-grained phase. For each \(GB_i\), the algorithm selects the two highest-degree nodes as centers and splits \(\mathcal{GB}_i\) into two new granular-balls, \(\mathcal{GB}_{c1}\) and \(\mathcal{GB}_{c2}\), based on the shortest path criterion:
\begin{equation}
c_1 = \mathop{\arg\max}_{v \in V_{{gb}_i}} \deg_{{gb}_i}(v), \quad c_2 = \mathop{\arg\max}_{v \in V_{{gb}_i} \setminus \{c_1\}} \deg_{{gb}_i}(v).
\end{equation}
The nodes in \(\mathcal{GB}_i\) are assigned to the nearest center \(c_1\) or \(c_2\) based on the shortest path:
\begin{equation}
v \in \mathcal{GB}_{c1} \iff d(v, c_1) \leq d(v, c_2),
\end{equation}
\begin{equation}
v \in \mathcal{GB}_{c2} \iff d(v, c_2) < d(v, c_1).
\end{equation}
The splitting continues iteratively until the purity of each child granular-ball meets a predefined threshold \(T = 1\), meaning that each granular-ball must contain nodes with the same label.  High purity \( T \) in a granular-ball indicates strong label consistency, preserving or enhancing the discriminative properties of the original graph. Conversely, low purity \( T \) suggests a heterogeneous label mix, potentially impairing model performance. Therefore, in preliminary experiments, the purity threshold is typically set high to encourage clear decision boundaries. This strict purity threshold ensures that the granular-balls are homogeneous in terms of node labels, which helps maintain label consistency and improves the performance of downstream classification tasks. Algorithm 3 details this process, dynamically adjusting granular-ball sizes to capture graph data complexity.

\subsubsection{Constructing the Granular-Ball Graph}

The construction of a granular-ball graph, \(\tilde{\mathcal{G}}\), involves a transformation from the set of granular- balls derived from the original graph \(\mathcal{G} = (\mathcal{V}, \mathcal{E}, \mathbf{X})\). Each granular-ball is represented as a super node in \(\tilde{\mathcal{G}}\), which initially contains no nodes or edges, i.e., \(\tilde{\mathcal{G}} = \emptyset\). Node indices from \(\mathcal{V}\) are associated with their respective granular-balls based on their labels and connectivity, defining a mapping \(f: \mathcal{V} \rightarrow \{\mathcal{GB}_1, \mathcal{GB}_2, \ldots, \mathcal{GB}_n\}\), where each \(\mathcal{GB}_i\) is a granular-ball. This mapping is used to establish edges between super nodes in \(\tilde{\mathcal{G}}\) as follows:
\begin{equation}
(u, v) \in \mathcal{E} \Rightarrow \left( f(u), f(v) \right) \in \tilde{\mathcal{E}} \text{ if } f(u) \neq f(v).
\end{equation}
This condition ensures that an edge is added between two different super nodes in \(\tilde{\mathcal{G}}\) if the corresponding original nodes \(u\) and \(v\) belong to different granular-balls. This connects granular-balls that share a direct link in the original graph, avoiding self-loops and ensuring that \(\tilde{\mathcal{G}}\) represents the higher-level structure of \(\mathcal{G}\). This methodology enhances the representation of graph by focusing on broader connectivity patterns while preserving essential graph properties.



\begin{algorithm}[tb!]
\caption{Training GNN with Granular-Ball Graph Coarsening}
\label{alg:training_gnn}
\raggedright
\begin{algorithmic}[1]
\Require Graph $\mathcal{G} = (\mathcal{V}, \mathcal{E}, \mathbf{X})$, Labels $\mathbf{Y}$
\Ensure Trained weight matrix \(\mathbf{W}^*\)

\State Apply the SGBGC algorithm on $\mathcal{G}$, and output a partition $\mathcal{P}$.
\State Construct the Granular-Ball Graph $\tilde{\mathcal{G}}$ using partition $\mathcal{P}$.
\State Compute the feature matrix of $\tilde{\mathcal{G}}$, $\tilde{\mathbf{X}} = \mathcal{P}^\top \mathbf{X}$.
\State Calculate the labels of $\tilde{\mathcal{G}}$, $\tilde{\mathbf{Y}} = \arg\max(\mathcal{P}\mathbf{Y})$.
\State Train parameter $\mathbf{W}$ to minimize the loss $\ell(\text{GNN}_{\tilde{\mathcal{G}}}(\mathbf{W}), \tilde{\mathbf{Y}})$.
\State Obtain an optimal weight matrix $\mathbf{W}^*$.
\State \Return $\mathbf{W}^*$
\end{algorithmic}
\end{algorithm}

\subsection{Training GNN with SGBGC}

Next, we incorporate the coarsened graph into the GNN models for training. We denote a GNN model based on graph $G$ as $\text{GNN}_{G}(\mathbf{W})$. Given a loss function $\ell$, such as cross-entropy, the loss of model is expressed as $\ell(\text{GNN}_{G}(\mathbf{W}), \mathbf{Y})$. We train a GNN model using $G_{\text{gb}}$ and the corresponding labels $\mathbf{\tilde{Y}}$, with the objective to minimize the loss function $\ell(\text{GNN}_{G_{\text{gb}}}(\mathbf{W}), \mathbf{\tilde{Y}})$ on the granular-ball coarsened graph, where $\mathbf{W}$ represents the model parameters. Ultimately, the optimal parameters $\mathbf{W}^*$ obtained from training on the coarsened graph are used in the GNN model on the original graph $G$ for downstream tasks.

In the granular-ball coarsened graph, each node is a super node, corresponding to a cluster of nodes in the original graph. The feature vector of each super node is the mean of all node feature vectors in the cluster, i.e., $\mathbf{\tilde{X}} = \mathbf{P}^\top \mathbf{X}$. We also set each label of super node accordingly, i.e., $\mathbf{\tilde{Y}} = \mathbf{P}^\top \mathbf{Y}$. However, super nodes may contain nodes from multiple categories. In such cases, we select the dominant label, i.e., the category with the highest purity, as the label for the super node by performing an argmax operation on $\mathbf{\tilde{Y}}$. During GCN model training, we can define the granular-ball graph convolution as:
\begin{equation}
\mathbf{Z}_{\text{gb}} = \mathbf{\tilde{D}}_{\text{gb}}^{-1/2} \mathbf{\tilde{A}}_{\text{gb}} \mathbf{\tilde{D}}_{\text{gb}}^{-1/2} \mathbf{\tilde{X}}_{\text{gb}} \mathbf{W}^*,
\end{equation}

where $\mathbf{Z}_{\text{gb}}$ denotes the output features of the granular-ball graph nodes, $\mathbf{\tilde{A}}_{\text{gb}} = \mathbf{\tilde{A}}_{\text{gb}} + \mathbf{\tilde{I}}_{\text{gb}}$, with $\mathbf{\tilde{A}}_{\text{gb}}$ being the adjacency matrix of the granular-ball graph and $\mathbf{\tilde{I}}_{\text{gb}}$ an identity matrix to incorporate self-connections. $\mathbf{\tilde{D}}_{\text{gb}}$ is the degree matrix of $\mathbf{\tilde{A}}_{\text{gb}}$, $\mathbf{\tilde{X}}_{\text{gb}}$ are the input features of the granular-ball graph nodes, and $\mathbf{W}_{\text{gb}}$ is the matrix of weights to be learned. Algorithm 1 details the use of granular-ball graph coarsening within the GNN framework.

\section{Experiments}
\subsection{Experimental Setup}

\subsubsection{Datasets}

We evaluate our methods on five standard node classification datasets: Cora, Citeseer, Pubmed\cite{9}, Co-CS, and Co-Phy\cite{48}. Each dataset features nodes represented as documents with sparse bag-of-words feature vectors, and edges representing citation links between these documents.

\subsubsection{Baselines}
To evaluate the effectiveness of SGBGC in graph coarsening, we compare our approach against three established baselines: FGC \cite{26}, SCAL \cite{21}, Gcond \cite{24}, and CMGC \cite{61}. FGC utilizes training to derive a coarsened graph, optimizing for feature and connection preservation. Gcond focuses on maintaining the spectral properties of the graph, beneficial for applications requiring eigenstructure preservation. CMGC aims to retain the graph convolution operations and improves the existing coarsening methods by merging structurally simlar nodes. SCAL, similar to SGBGC, pre-processes the graph to produce a coarsened version by incorporating various  coarsening techniques, specifically VNGC, VEGC\cite{28}, GSGC, and JCGC\cite{51}.

\subsubsection{Implementation Details}
All models were implemented using Python and PyTorch Geometric, with experiments conducted on an Intel(R) Xeon(R) W-2245 CPU @3.90GHz and an NVIDIA GeForce RTX 3090 GPU. The dataset splits were handled according to the methodologies described in references \cite{21}, and accuracy (ACC) was used as the evaluation metric. Results reported are averages over 20 runs, along with standard deviations to indicate variability. To ensure fairness in comparisons, our models utilized the same network architectures as the baselines. We uniformly use Adam optimizer with learning rates of 0.01 and set the weight decay to 0.0005. For the coarse GCN, the number of training epochs are 200 and the number of hidden laywers are 64 and the early stopping is set to 10.
\subsection{Experiment Results}

\begin{table*}[tb!]
\small
\setlength{\tabcolsep}{4pt}
\caption{Performance comparison of SGBGC and other coarsening methods under different ratios (r) for node classification in GNNs.(mean±std\%, the best results are bolded, the second results are underlined, and "-" means out of memory).}
\centering
\begin{tabular}{lccccccccc}
\toprule
\textbf{DataSet} & \textbf{Ratio (r)} & \textbf{VNGC} & \textbf{VEGC} & \textbf{JCGC} & \textbf{GSGC} & \textbf{GCOND} & \textbf{FGC} & \textbf{CMGC}  &  \textbf{SGBGC} \\
\midrule
\multirow{3}{*}{Cora} & 0.5 & 77.28$\pm$1.18 & 78.36$\pm$1.75 & 79.85$\pm$1.24 & 81.31$\pm$1.59 &  - & 86.22$\pm$0.15 & \textbf{88.40$\pm$0.67}  & \underline{87.85$\pm$0.64} \\
 & 0.3 & 78.33$\pm$1.78 & 76.97$\pm$1.86 & 77.79$\pm$1.84 & 80.05$\pm$0.86& 76.49$\pm$0.71 & 85.39$\pm$0.27 & \underline{85.96$\pm$0.84} & \textbf{86.29$\pm$0.79} \\
 & 0.1 & 69.19$\pm$0.73 & 62.76$\pm$1.94 & 67.44$\pm$1.25 & 68.32$\pm$1.08 & 80.54$\pm$1.61 & 81.66$\pm$0.36 & \underline{82.50$\pm$0.93} & \textbf{82.99$\pm$1.00} \\
\midrule
\multirow{3}{*}{Citeseer} & 0.5 & 73.16$\pm$0.64 & 71.31$\pm$1.10 & 71.91$\pm$1.56 & 71.23$\pm$1.69 & - & \underline{76.28$\pm$0.21} & \textbf{77.33$\pm$0.85} & 75.25$\pm$1.02 \\
 & 0.3 & 70.12$\pm$1.16 & 42.32$\pm$2.08 & 67.58$\pm$1.12 & 66.13$\pm$2.03 & 71.26$\pm$3.01 & 73.16$\pm$0.27 & \underline{74.85$\pm$0.92} & \textbf{75.73$\pm$0.88} \\
 & 0.1 & 60.97$\pm$2.79 & 41.88$\pm$1.18 & 52.68$\pm$2.35 & 55.24$\pm$1.52 & 69.67$\pm$2.37 & 72.87$\pm$1.13 & \underline{73.52$\pm$1.01} & \textbf{73.69$\pm$0.93} \\
\midrule
\multirow{3}{*}{Pubmed} & 0.3 & 83.95$\pm$0.21 & 82.54$\pm$0.35 & 82.13$\pm$0.64 & 83.99$\pm$0.27  & - & 83.36$\pm$0.05 & \underline{85.80$\pm$0.30} & \textbf{86.28$\pm$0.20} \\
 & 0.1 & 81.55$\pm$0.28 & 82.49$\pm$0.50 & 81.83$\pm$0.54 & 82.96$\pm$0.34 & - & 82.44$\pm$0.07 & \underline{83.75$\pm$0.45} & \textbf{84.04$\pm$0.36} \\
 & 0.05 & 78.12$\pm$3.00 & \underline{82.48$\pm$0.42} & 81.51$\pm$1.31 & 81.81$\pm$0.31 & 80.23$\pm$0.08 & 81.64$\pm$0.29 & 82.37$\pm$0.53 & \textbf{82.50$\pm$0.38} \\
\midrule
\multirow{3}{*}{Co-CS} & 0.3 & 56.05$\pm$5.36 & 53.61$\pm$4.19 & 46.87$\pm$4.59 & 52.76$\pm$4.78 & - & \underline{89.94$\pm$0.04} & 75.54$\pm$2.35 & \textbf{93.09$\pm$0.13} \\
 & 0.1 & 37.67$\pm$2.62 & 33.69$\pm$2.96 & 33.14$\pm$0.36 & 39.79$\pm$0.96 & -& \underline{88.47$\pm$0.06} & 72.57$\pm$3.04  & \textbf{92.22$\pm$0.22} \\
 & 0.05 & 23.22$\pm$0.12 & 34.05$\pm$1.16 & 27.09$\pm$3.46 & 33.08$\pm$0.41 & 82.76$\pm$0.94 & \underline{83.85$\pm$1.35} & 83.34$\pm$1.87 & \textbf{91.21$\pm$0.30} \\
\midrule
\multirow{3}{*}{Co-Phy} & 0.3 & 90.76$\pm$2.04 & 90.66$\pm$2.77 & 86.33$\pm$2.79 & 89.02$\pm$3.87 & - & 89.94$\pm$0.15 & \underline{96.30$\pm$0.13} & \textbf{96.35$\pm$0.09} \\
 & 0.1 & 76.00$\pm$0.94 & 77.73$\pm$2.29 & 73.61$\pm$2.41 & 69.86$\pm$0.09 & - & 89.23$\pm$0.28 & \underline{95.76$\pm$0.16} & \textbf{95.88$\pm$0.11} \\
 & 0.05 & 72.18$\pm$1.76 & 76.99$\pm$1.90 & 72.42$\pm$1.50 & 69.84$\pm$0.15 & 85.06$\pm$0.33 & 81.91$\pm$0.15 & \underline{95.36$\pm$0.21} & \textbf{95.61$\pm$0.14} \\
\bottomrule
\end{tabular}
\label{table:results}
\end{table*}

\begin{figure}[tb!]
  \centering
  \includegraphics[width=\linewidth]{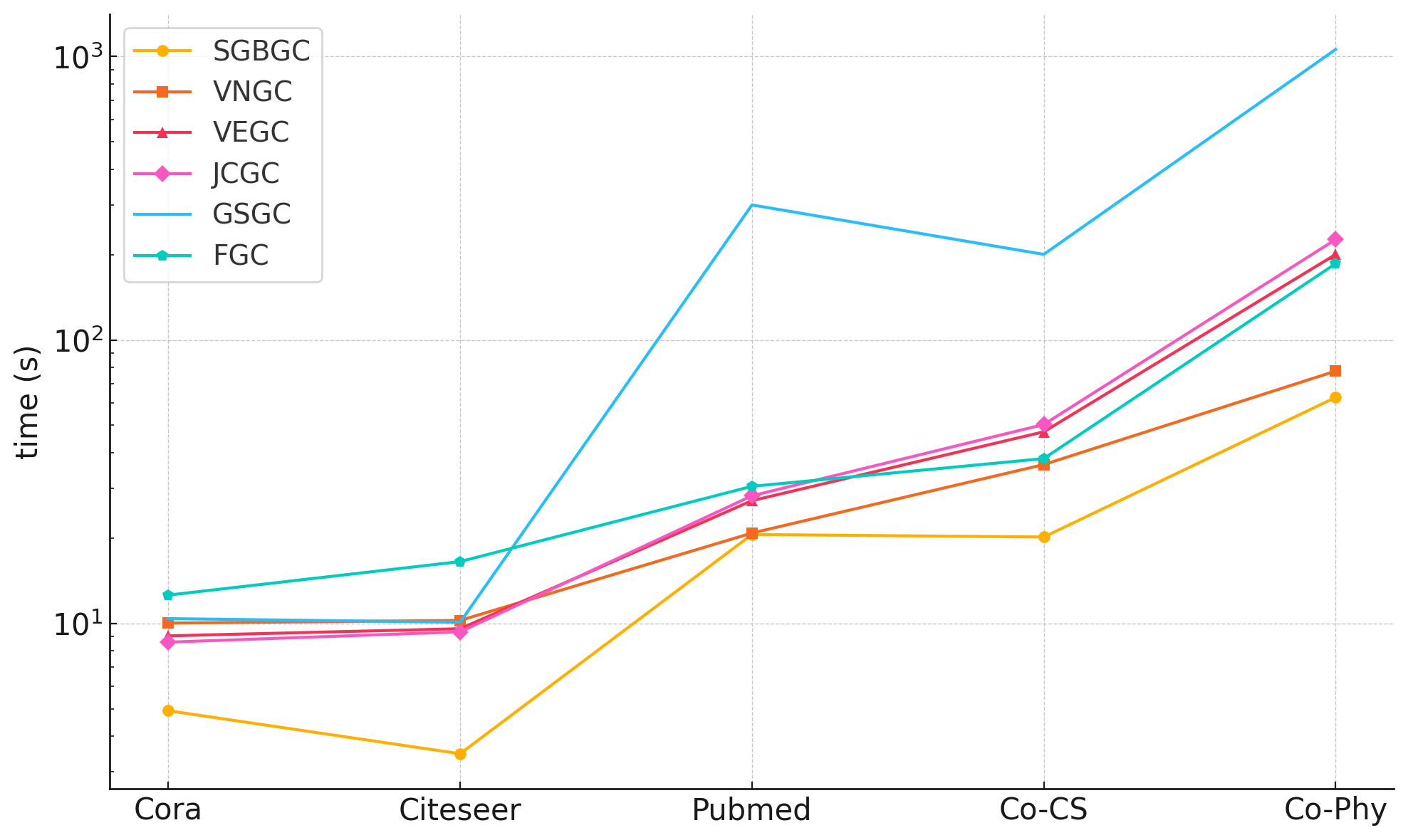}
  \caption{Comparison of time costs among different coarsening methods.}
  \label{fig:time-comparison}
\end{figure}

\begin{figure}[tb!]
  \centering
  \includegraphics[width=0.9\columnwidth]{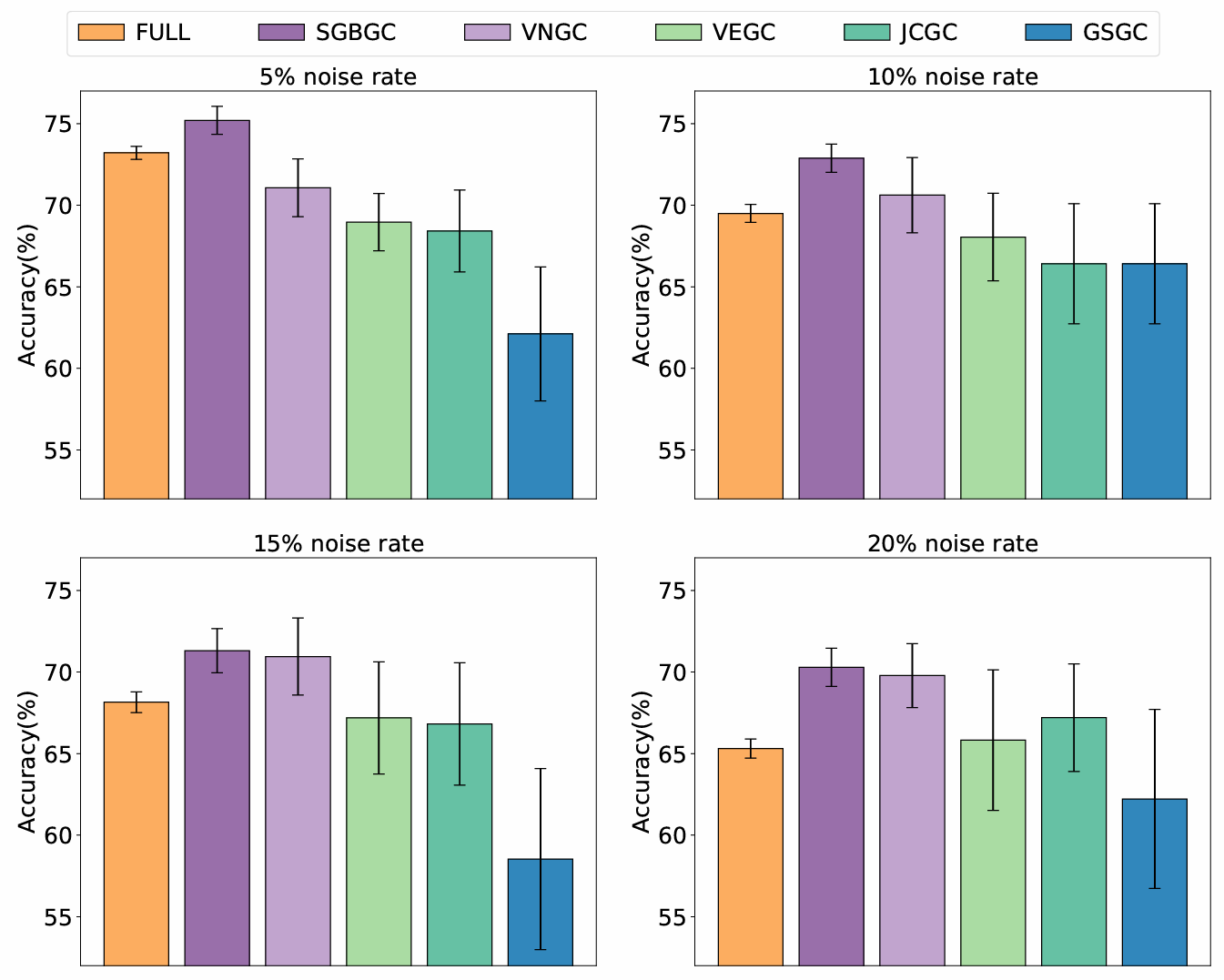}
  \caption{Comparison of different noise rates on Citeseer with 0.3 coarsening ratio, including original graph and various coarsening methods.}
  \label{fig:noise-exp}
\end{figure}

\begin{figure*}[tb!]
  \centering
  \includegraphics[width=1\textwidth]{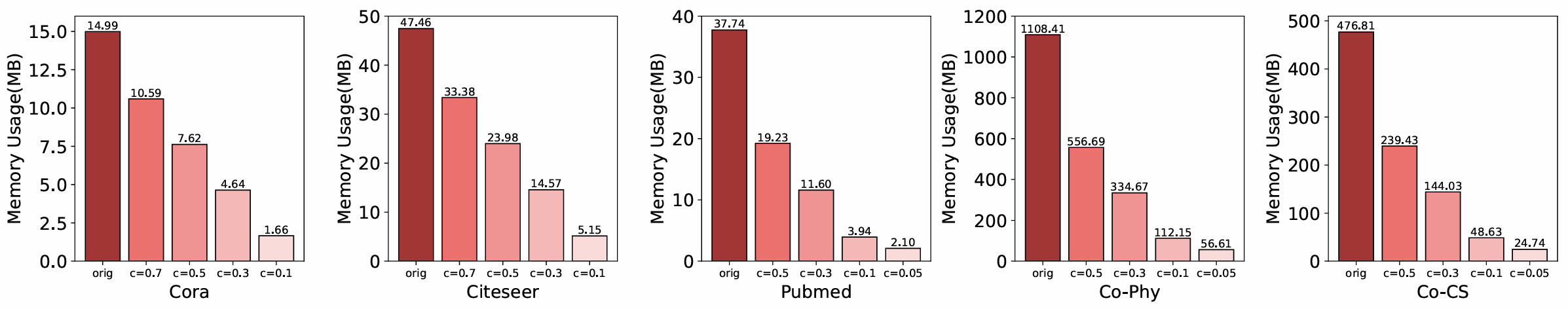}
  \caption{The memory usage of APPNP and granular-ball coarsened APPNP.}
  \label{fig:memory}
\end{figure*}

\subsubsection{Node Classification Performance}

In a transductive setting, using GCN as the base model, as shown in Table 1, it is evident that our SGBGC method not only maintains accuracy comparable to the original graph but also achieves significant improvements over existing unsupervised graph coarsening methods. The GCOND, FGC, and CMGC methods require training for coarsening, which is highly time-consuming. Additionally, GCOND often runs out of memory capacity at higher coarsening rates. In contrast, SGBGC is less time-intensive while still ensuring accuracy across most coarsening rate settings. Notably, at a coarsening rate of 0.1, our method demonstrates a substantial performance boost, achieving an accuracy of 82.99\%, which is markedly higher than that of other methods. This indicates that even under extreme graph compression, our method preserves high accuracy, showcasing robust information retention capabilities. In the Citeseer dataset, our method demonstrates superior performance across most tested coarsening rates. Specifically, at a coarsening rate of 0.1, our method achieved an accuracy of 73.6\%, which represents a significant improvement over other methods. Although at a coarsening rate of 0.5 our method is slightly outperformed by CMGC, the overall results still confirm the generalizability and robustness of our approach across different graph datasets. Results from the Pubmed dataset also support the superiority of our method, particularly at a coarsening rate of 0.3, where the accuracy reached 86.28\%, showing minimal performance loss compared to the original graph. On the Co-CS and Co-Phy datasets, our method exhibited remarkable performance improvements. Especially in the Co-CS dataset, even at very low coarsening rates (such as 0.1 and 0.05), the accuracy of SGBGC exceeded 91.21\%, far surpassing other methods. These results emphasize the powerful capability of our method in handling large-scale graph data with complex structures. By dynamically leveraging label information and a purity-based granular-ball splitting strategy, our method effectively enhances the quality of graph coarsening while also ensuring efficient computational performance.

\begin{table}[tb!]
\small
\centering
\caption{Comparison of classification performance with adaptive granular-ball coarsening rates. FULL represents performance on uncoarsened datasets.}
\label{tab:classification_performance}
\begin{tabular}{@{}lcccccc@{}}
\toprule
Datasets & Cora & Citeseer & Pubmed & Co-CS & Co-Phy \\ 
\midrule
Ratio (r) & 0.45 & 0.49 & 0.42 & 0.41 & 0.36 \\
\midrule
\textbf{SGBGC} & \textbf{87.57} & \textbf{75.38} & \textbf{86.29} & \textbf{93.28} & \textbf{96.39} \\
FULL & 88.38 & 76.48 & 86.40 & 93.78 & 96.65 \\
\bottomrule
\end{tabular}
\end{table}

\subsubsection{Adaptive approach}

Existing graph coarsening methods often lack support for adaptive coarsening, requiring a predetermined coarsening rate to proceed. This limitation can hinder their flexibility and scalability. In contrast, our SGBGC method introduces an adaptive coarsening mechanism, where the coarsening rate is not pre-specified but is instead determined dynamically through granular-ball splitting. This allows SGBGC to adjust autonomously, enhancing scalability and adaptability to various graph sizes. The results in Table 2 demonstrate that under adaptive conditions, SGBGC achieves performance very close to that of the original graph, validating the effectiveness of our approach.

\subsubsection{Computational Efficiency of SGBGC}

For a fair comparison of coarsening methods, we set the coarsening rates as follows: 0.3 for the Cora and Citeseer datasets, and 0.05 for the PubMed, Co-CS, and Co-Phy datasets. Figure 3 shows the performance results. On the Cora dataset, SGBGC required only 4.92 seconds, which is 51\% faster than VNGC and 46\% faster than VEGC. On the Citeseer dataset, SGBGC completed processing in 3.47 seconds, which is 66\% faster than both VNGC and GSGC.

For larger datasets, the advantages of SGBGC become more pronounced. On the Co-CS dataset, SGBGC completed processing in 20.18 seconds, outperforming VNGC by 44\% and GSGC by over 9 times. Similarly, on the Co-Phy dataset, SGBGC finished in 62.72 seconds, nearly 17 times faster than GSGC. These results clearly indicate that the SGBGC method consistently outperforms other coarsening methods in terms of speed, especially on larger datasets. The superior performance of SGBGC is likely due to its reliance on local computations (i.e., granular-ball subgraphs) rather than global computations, making it highly efficient for large-scale datasets. Due to its extremely slow processing time, Gcond and CMGC were excluded from our comparisons; for example, Gcond took nearly two hours and CMGC over a thousand seconds on the Cora dataset. 

\begin{figure}[tb!]
  \centering
  \includegraphics[width=1\linewidth]{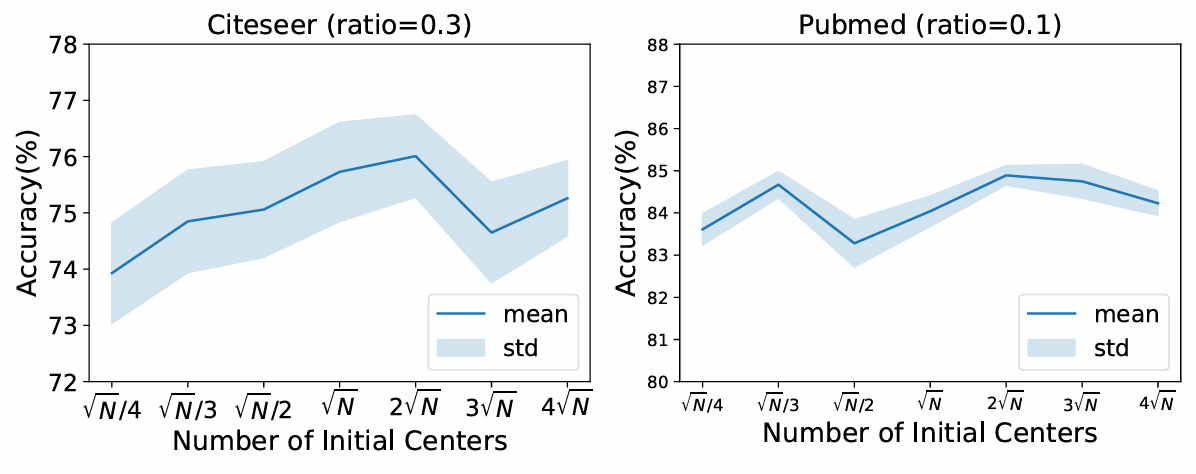}
  \caption{Parametric Analysis of Initial Centers}
  \label{fig:Parametric Analysis}
\end{figure}

\subsubsection{Memory Usage}

Figure 5 shows the memory usage of APPNP at different coarsening ratios. Since parameter space is negligible compared to input tensor size, coarse APPNP's memory usage is roughly proportional to the coarsening ratio. Our method significantly reduces memory consumption by several orders of magnitude on medium to large datasets.

\subsubsection{Robustness Analysis Under Different Noise Rates}

Figure 4 evaluates various coarsening methods under 5\%, 10\%, 15\%, and 20\% noise on the Citeseer dataset with a 0.3 coarsening ratio. SGBGC consistently outperformed other methods and the noisy original graph. At 5\% noise, SGBGC achieved 75.20\% accuracy, surpassing the original graph's 73.21\%. Even at higher noise levels, SGBGC maintained superior accuracy, significantly better than both the original graph and the next best method, VNGC. SGBGC's robustness likely stems from granular-ball computing, which mitigates label noise through node clustering based on purity thresholds. This approach preserves structural and feature integrity, enhancing node classification accuracy. These results underscore SGBGC's superior noise tolerance, making it a robust choice for real-world applications.

\subsubsection{Parameter Analysis}

We experimented with different initial center numbers, as shown in Figure 6. The results indicate that our method maintains consistent performance across varying settings, demonstrating strong stability. Experimental results show that in most cases, when the number of initial centers is too large, the accuracy will be slightly improved, but the efficiency will be correspondingly slower. For example, on the Pubmed dataset, coarsening takes 29.7 seconds with \(\sqrt{N}\) centers, 35.0 seconds with 2\(\sqrt{N}\), and 78.4 seconds with 4\(\sqrt{N}\), with only minor accuracy gains. Thus, \(\sqrt{N}\) was chosen as the optimal value, informed by both our experiments and empirical evidence from the previous work \cite{50}, which has been shown to balance the trade-off between computational complexity and accuracy.

\section{Conclusion and Future Work}
\label{Conclusion}
In this work, we introduce Supervised Granular-Ball Graph Coarsening (SGBGC), a novel method for supervised graph coarsening that significantly enhances the scalability of GNNs by compressing the original graph over tenfold without losing accuracy. SGBGC is an efficient and adaptive approach requiring minimal hyperparameters, and it can also operate in a non-adaptive form based on the coarsening ratio. Experiments show that SGBGC outperforms several advanced methods in both precision and efficiency, notably accelerating GNN training in  node classification tasks. While our current architecture is focused on node classification as a downstream task, future work will explore other applications such as graph classification and link prediction to further broaden the applicability of our graph coarsening method.

\section{Acknowledgments}
We sincerely appreciate the dedication and efforts of everyone who contributed to this work. Thank you all for your hard work. This work was supported in part by the National Natural Science Foundation of China under Grant Nos. 62222601, 62450043, 62176033, 62221005 and 61936001, key cooperation project of Chongqing municipal education commission No.HZ2021008.

\appendix
\bibliography{bibtex}

\appendix

\newpage

\section{Related Works}
\label{appendix:a}

\subsection{Graph Neural Networks}

Graph Neural Networks (GNNs) \cite{9,10,11,12} adhere to a message-passing paradigm, whereby the representation of a node is iteratively updated through aggregating representations from neighboring nodes. When generalized to large graphs, GNNs face similar scalability issues, primarily due to the uncontrollable expansion of neighborhoods during the aggregation phase. A prominent approach is to decouple the interdependencies among nodes, thereby reducing the receptive field. The hierarchical sampling combined with mini-batch training, pioneered by \cite{10}, has proven to be a highly effective strategy. Since then, several subsequent works have sought to enhance this baseline by optimizing the sampling process, improving random estimations, and other extensions \cite{16,13,15,14}.

\subsection{Noisy Labels in Graph Neural Networks}
\label{appendix:a.4}

Deep neural networks are known to overfit to noisy labels, resulting in poor generalization performance, as shown in \cite{58}. This issue is particularly pertinent in the context of Graph Neural Networks (GNNs), which are also susceptible to the influence of noisy labels \cite{56,57}. Due to the message-passing mechanism inherent in GNNs, noisy label information can propagate to unlabelled nodes, significantly degrading performance of the model.

To address this, several methods have been proposed. For instance, \cite{57} introduced D-GNN, which applies backward loss correction to mitigate the effects of noisy labels. Similarly, Zhang et al. \cite{56} proposed adding a regularization term that encourages the learned representations to accurately predict community labels, thereby avoiding overfitting to noisy labels. Furthermore, \cite{55} proposed a strategy to alleviate the negative impact of label noise by connecting unlabelled nodes with labelled nodes that have high feature similarity. This approach introduces cleaner label information and generates accurate pseudo-labels, providing additional supervision and further reducing the impact of noisy labels.

\newtheorem{definition}{Definition}
\newtheorem{theorem}{Theorem}

\section{Theoretical Foundations}
\label{appendix:B}

\subsection{The Details of Iterative Granular-Ball Splitting}

The granular-ball splitting process is dynamically quantified through iterative updates of the projection matrix \(\mathbf{P}\) and granular-ball subgraphs. This approach ensures a rigorous and adaptive method to capture the evolving structure of the graph. Initially, the projection matrix \(\mathbf{P}^{(0)} \in \mathbb{R}^{N \times \gamma}\) is defined as:
\begin{equation}
\mathbf{P}_{ij}^{(0)} = 
\begin{cases} 
1 & \text{if node } v_i \text{ belongs to granular ball } \mathcal{GB}_j, \\
0 & \text{otherwise}
\end{cases}
\end{equation}

The initial granular-ball subgraphs are defined as \(\mathcal{GB}_i^{(0)} = (\mathcal{V}_{gb}^{(0)}, \mathcal{E}_{gb}^{(0)}, \mathbf{X}_{gb}^{(0)})\).

In each iteration \(t\), we update the projection matrix \(\mathbf{P}^{(t)}\) and granular-ball subgraphs \(\mathcal{GB}_i^{(t)}\). For each granular ball \(\mathcal{GB}_i^{(t)}\), select the two highest-degree nodes \(c_1\) and \(c_2\) as new centers:
\begin{align}
c_1 &= \mathop{\arg\max}_{v \in \mathcal{V}_{\mathcal{GB}_i^{(t)}}} \deg(v), \\
c_2 &= \mathop{\arg\max}_{v \in \mathcal{V}_{\mathcal{GB}_i^{(t)}} \setminus \{c_1\}} \deg(v)
\end{align}

Reassign nodes based on the shortest path criterion:
\begin{align}
v &\in \mathcal{GB}_{c1}^{(t+1)} \iff d(v, c_1) \leq d(v, c_2), \\
v &\in \mathcal{GB}_{c2}^{(t+1)} \iff d(v, c_2) < d(v, c_1)
\end{align}

Update the projection matrix \(\mathbf{P}^{(t+1)}\):
\begin{equation}
\mathbf{P}_{ij}^{(t+1)} = 
\begin{cases} 
1 & \text{if node } v_i \text{ belongs to new granular ball } \mathcal{GB}_j, \\
0 & \text{otherwise}
\end{cases}
\end{equation}

With each update of the projection matrix, the granular-ball subgraphs \(\mathcal{GB}_i^{(t+1)} = (\mathcal{V}_{gb}^{(t+1)}, \mathcal{E}_{gb}^{(t+1)}, \mathbf{X}_{gb}^{(t+1)})\) also change:
\begin{align}
\mathcal{V}_{gb}^{(t+1)} &= \{v \in \mathcal{V} : \mathbf{P}_{vj}^{(t+1)} = 1 \text{ for some } j\}, \\
\mathcal{E}_{gb}^{(t+1)} &= \{(u, v) \in \mathcal{E} : u, v \in \mathcal{V}_{gb}^{(t+1)}\}, \\
\mathbf{X}_{gb}^{(t+1)} &= \{\mathbf{x}_v : v \in \mathcal{V}_{gb}^{(t+1)}\}.
\end{align}
Iterate until all granular-ball subgraphs achieve a purity of \( T = 1 \), meaning that all \( v \in \mathcal{V}_{gb}^{(t+1)} \) within each \(\mathcal{GB}_i^{(t+1)}\) have the same label. This strict purity threshold ensures that the granular-balls are homogeneous in terms of node labels, which helps maintain label consistency and improves the performance of downstream classification tasks.

\subsection{Theoretical Analysis for SGBGC}

Define the adjacency matrix \(\mathbf{A}\) of the original graph \(\mathcal{G} = (\mathcal{V}, \mathcal{E}, \mathbf{X})\) and the degree matrix \(\mathbf{D}\). The Laplacian matrix \(\mathbf{L}\) is given by:
\begin{equation}
\mathbf{L} = \mathbf{D} - \mathbf{A}.
\end{equation}
Next, construct the coarsened graph \(\mathbf{G}_{\text{gb}}\) by creating a projection matrix \(\mathbf{P}\) that maps nodes in \(\mathbf{V}\) to nodes in \(\tilde{\mathbf{V}}\). The projection matrix \(\mathbf{P}\) is defined as:
\begin{equation}
\mathbf{P} \in \mathbb{R}^{N \times n},
\end{equation}
where \(\mathbf{P}[i, j] = 1\) if node \(i\) in the original graph is mapped to super node \(j\) in the coarsened graph. To ensure label consistency, use label information \(\mathbf{Y}\) so that nodes with the same label are grouped together into the same granular ball. Adjust the coarsening process by incorporating a label matrix \(\mathbf{Y}\) into the projection matrix \(\mathbf{P}\). The Laplacian matrix of the coarsened graph \(\mathbf{L}_{\text{gb}}\) is computed using the projection matrix:
\begin{equation}
\mathbf{L}_{\text{gb}} = \mathbf{P}^{\top} \mathbf{L} \mathbf{P}.
\end{equation}
This computation ensures that the spectral properties of the original graph are preserved in the coarsened graph. Assume we have an original graph \( \mathcal{G} = (\mathcal{V}, \mathcal{E}, \mathbf{X})\), where \( |\mathcal{V}| = N \) nodes and \( |\mathcal{E}| = M \) edges. We generate a coarsened graph \( \mathcal{G}_{\text{gb}} = (\tilde{\mathcal{V}}, \tilde{\mathcal{E}}, \tilde{\mathbf{X}})\) 
using the granular-ball method, where \( |\tilde{\mathcal{V}}| = n \) super nodes, and the initial number of granular-balls is \( \gamma = \sqrt{N} \).

\setcounter{definition}{2}
\begin{definition}[Label Consistency Measure]
We introduce a label consistency measure \( C(y) \) defined as:
\begin{equation}
C(y) = \sum_{v_i, v_j \in GB} \mathbb{I}(y_i = y_j),
\end{equation}
where \( \mathbb{I} \) is the indicator function that is 1 if the condition is satisfied, otherwise 0.
\end{definition}

\begin{definition}[Rayleigh Quotient for Granular-Ball Graph]
The Rayleigh quotient \( R(\mathbf{x}) \) for the original graph is defined as:
\begin{equation}
R(\mathbf{x}) = \frac{\mathbf{x}^T \mathbf{L} \mathbf{x}}{\mathbf{x}^T \mathbf{x}},
\end{equation}

To incorporate label consistency, we define a new Rayleigh quotient \( R_C(\mathbf{x}) \) as:
\begin{equation}
R_C(\mathbf{x}) = \frac{\mathbf{x}^T \mathbf{L} \mathbf{x} + \lambda C(y)}{\mathbf{x}^T \mathbf{x}},
\end{equation}

During the granular-ball process, we construct a granular-ball graph \( \mathbf{G}_{\text{gb}} \) and define its projection matrix \( \mathbf{P} \). The Rayleigh quotient for the granular-ball graph \( R_{\text{gb}}(\tilde{\mathbf{x}}) \) is then given by:
\begin{equation}
R_{\text{gb}}(\tilde{\mathbf{x}}) = \frac{(\mathbf{P}^T \mathbf{x})^T \mathbf{L}_{\text{gb}} (\mathbf{P}^T \mathbf{x}) + \lambda C(\tilde{y})}{(\mathbf{P}^T \mathbf{x})^T (\mathbf{P}^T \mathbf{x})}.
\end{equation}
\end{definition}

\begin{theorem}
Assuming \( \mathbf{P} \) is a generalized orthogonal matrix (\( \mathbf{P}^T \mathbf{P} \approx \mathbf{I} \)), we have:
\begin{equation}
R_{\text{gb}}(\tilde{\mathbf{x}}) \approx \frac{\mathbf{x}^T \mathbf{L} \mathbf{x} + \lambda C(y)}{\mathbf{x}^T \mathbf{x}} = R_C(\mathbf{x}).
\end{equation}
\end{theorem}

This shows that when the label consistency measure \( C(y) \) approaches its maximum, the Rayleigh quotient of the granular-ball graph is approximately equal to the Rayleigh quotient of the original graph, thereby preserving the spectral properties of the original graph.

\begin{proof}
Considering the assumption that \(\mathbf{P}\) is a generalized orthogonal matrix, i.e., \(\mathbf{P}^T \mathbf{P} \approx \mathbf{I}\), we derive the transformation of the numerator and denominator as follows.

We start with the denominator:
\begin{equation}
(\mathbf{P}^T \mathbf{x})^T (\mathbf{P}^T \mathbf{x}),
\end{equation}

Using the orthogonality of \(\mathbf{P}\):
\begin{equation}
(\mathbf{P}^T \mathbf{x})^T (\mathbf{P}^T \mathbf{x}) \approx \mathbf{x}^T \mathbf{P} (\mathbf{P}^T \mathbf{x}) = \mathbf{x}^T \mathbf{I} \mathbf{x} = \mathbf{x}^T \mathbf{x},
\end{equation}

Next, we consider the numerator:
\begin{equation}
(\mathbf{P}^T \mathbf{x})^T \mathbf{L}_{\text{gb}} (\mathbf{P}^T \mathbf{x}) + \lambda C(\tilde{y}).
\end{equation}

First, we approximate the quadratic form:
\begin{equation}
(\mathbf{P}^T \mathbf{x})^T \mathbf{L}_{\text{gb}} (\mathbf{P}^T \mathbf{x}) \approx \mathbf{x}^T \mathbf{P} \mathbf{L}_{\text{gb}} \mathbf{P}^T \mathbf{x},
\end{equation}

Assuming \(\mathbf{L}_{\text{gb}} \approx \mathbf{L}\), we get:
\begin{equation}
\mathbf{P} \mathbf{L}_{\text{gb}} \mathbf{P}^T \approx \mathbf{L},
\end{equation}

Thus:
\begin{equation}
\mathbf{x}^T \mathbf{P} \mathbf{L}_{\text{gb}} \mathbf{P}^T \mathbf{x} \approx \mathbf{x}^T \mathbf{L} \mathbf{x}.
\end{equation}

Next, for the label consistency term, we assume:
\begin{equation}
\lambda C(\tilde{y}) \approx \lambda C(y),
\end{equation}

Combining these approximations, the numerator becomes:
\begin{equation}
(\mathbf{P}^T \mathbf{x})^T \mathbf{L}_{\text{gb}} (\mathbf{P}^T \mathbf{x}) + \lambda C(\tilde{y}) \approx \mathbf{x}^T \mathbf{L} \mathbf{x} + \lambda C(y).
\end{equation}

Finally, using these approximations, the Rayleigh quotient for the granular-ball graph \(R_{\text{gb}}(\tilde{\mathbf{x}})\) can be approximated as:
\begin{equation}
R_{\text{gb}}(\tilde{\mathbf{x}}) \approx \frac{\mathbf{x}^T \mathbf{L} \mathbf{x} + \lambda C(y)}{\mathbf{x}^T \mathbf{x}} = R_C(\mathbf{x}).
\end{equation}

\end{proof}

\section{Computational complexity}
\label{fig:Computational complexity}

The time complexity of the Supervised Granular-Ball Graph Coarsening (SGBGC) method is derived as follows: Let the number of nodes and edges in the original graph be \(N\) and \(M\), respectively.

\textbf{Initial Granular-Ball Formation Phase:} Identifying the node with the maximum degree requires sorting, which has a time complexity of \(O(N \log N)\). Next, finding \(\sqrt{N}\) central nodes involves calculating the single-source shortest path \(\sqrt{N}\) times. The time complexity of finding a single-source shortest path in an unweighted graph using BFS is \(O(N+M)\). Since this needs to be performed for each central node, the time complexity for node allocation is \(O(\sqrt{N}(N+M))\). Thus, the total time complexity for the coarse partitioning phase is \(O(\sqrt{N}(N+M) + N \log N)\).

\textbf{Fine Granular-Ball Bifurcation:} Let the number of nodes and edges in each granular-ball be n and m, respectively. Sorting to find the node with the maximum degree has a time complexity of \(O(n \log n)\). Calculating the shortest path distribution to two central nodes for each non-central node using the BFS algorithm has a time complexity of \(O(n+m)\). As there are two central nodes, the total time complexity is \(O(2(n+m))\). Therefore, the total time complexity of the bifurcation stage is \(O(n \log n + 2(n+m))\).

\textbf{Construction of the Granular-Ball Graph:} Let the number of granular-balls be L. Adding each granular-ball to the graph has a time complexity of \(O(L)\). Constructing a map from the original nodes to the granular-ball index has a time complexity of \(O(N)\). Constructing edges between granular-balls has a time complexity of \(O(M)\). Therefore, the total time complexity of the construction stage is \(O(L+N+M)\).

Summarizing, the overall time complexity of SGBGC is \(O(N^{3/2} + M\sqrt{N} + N \log N + n \log n + 2(n+m) + L + N + M)\). Given that \(n \ll N\), \(m \ll M\) and \(L \ll N\), the total time complexity of SGBGC is \(O(N^{3/2} + M\sqrt{N})\).

Unlike traditional graph coarsening techniques, SGBGC employs a multi-granularity strategy to optimize node processing, thereby enhancing computational efficiency. SGBGC does not require pre-computation of distances between all fine-grained node pairs in the graph. Instead, it dynamically processes nodes of different granular-ball sizes, adapting to the data distribution and fitting most data scenarios. This approach offers greater generalization capability and higher time efficiency.

\section{Datasets and Training details}
\subsection{Datasets}
For experiments on transductive node classification, we evaluate our model on five citation network datasets: Cora, Citeseer, Pubmed \cite{9}, Co-CS, and Co-Phy \cite{48}. These datasets contain sparse bag-of-words feature vectors for each document and a list of citation links between documents. The citation links are treated as undirected edges, and each document has a class label. These datasets provide a comprehensive benchmark for evaluating the scalability and performance of our model. The complete information of these datasets is shown in Table \ref{tab:datasets}.

\begin{table}[ht]
\centering
\caption{Summary of Datasets Used in Experiments}
\label{tab:datasets}
\begin{tabular}{lcccc}
\toprule
\textbf{Dataset} & \textbf{\#Nodes} & \textbf{\#Edges} & \textbf{\#Features} & \textbf{\#Classes} \\
\midrule
Cora     & 2,708  & 5,429   & 1,433  & 7 \\
Citeseer & 3,327  & 4,732   & 3,703  & 6 \\
Pubmed   & 19,717 & 44,338  & 500    & 3 \\
Co-CS    & 18,333 & 182,121 & 6,805  & 15 \\
Co-phy   & 34,493 & 247,962 & 8,415  & 5 \\
\bottomrule
\end{tabular}
\end{table}

\subsection{Training details}
\label{tab:Training details}
Our experiments primarily follow a full-supervised node classification setup, as referenced in \cite{49,59}, with datasets split using a random split of 60\%/20\%/20\% for training, validation, and testing. Given that our coarsening method relies heavily on label information, full-supervised node classification is essential when sufficient labeled nodes are available. However, to evaluate the robustness and generalizability of our method, we have also incorporated semi-supervised settings in our experiments. In these semi-supervised setups, only a fraction of the labeled nodes are used for training, which tests the effectiveness of our approach in scenarios where label information is limited.

To ensure reproducibility, we provide detailed experimental procedures. During the coarsening process, unlabelled isolated nodes were removed from the training and validation sets. We employed accuracy (ACC) and computational time as metrics to evaluate the performance of the graph coarsening methods. Following the SCAL experimental framework \cite{21}, we used GCN, GAT, and APPNP as benchmark models. For fairness in comparisons, methods including VNGC, VEGC, JCGC, GSGC, and FGC followed the adaptive coarsening ratio \( r \) achieved by SGBGC across various datasets. However, GCond and FGC rely on training and thus did not use this split method. To ensure generalizability and mitigate evaluation bias due to randomness, we introduced multiple coarsening levels and averaged results over twenty runs for the SGBGC method. The accuracy for the GCond method was determined by averaging the outcomes of ten independent runs without adjusting for the coarsening ratio \( r \). This approach enhances the robustness and reliability of our performance assessments by reducing the impact of outliers.

The coarsening ratio \( r \), defined as the ratio of the number of nodes in the coarsened graph to that in the original graph, ranges between 0 and 1. The method yielding the highest accuracy on each dataset is highlighted in bold, as summarized in Table \ref{tab:classification_performance}. For implementation specifics, we adhered to the hyperparameter settings recommended in prior works \cite{47,48} for the conventional GCN, GAT, and APPNP architectures.

\section{More Experiments}
\label{fig:More Experiments}

\subsection{Applicable to a General GNN Framework}

Our proposed Supervised Granular-Ball Graph Coarsening (SGBGC) method is a versatile preprocessing technique that can be seamlessly integrated into various Graph Neural Network (GNN) architectures, such as GCN, APPNP, and GAT. Table \ref{tab:classification_performance} presents the performance of these GCN on several datasets at specified coarsening ratios. For Cora and Citeseer, the coarsening ratio was set to 0.3, while for Pubmed, Co-CS, and Co-Phy, the coarsening ratio was set to 0.05.

\begin{figure}
  \centering
  \includegraphics[width=0.48\textwidth]{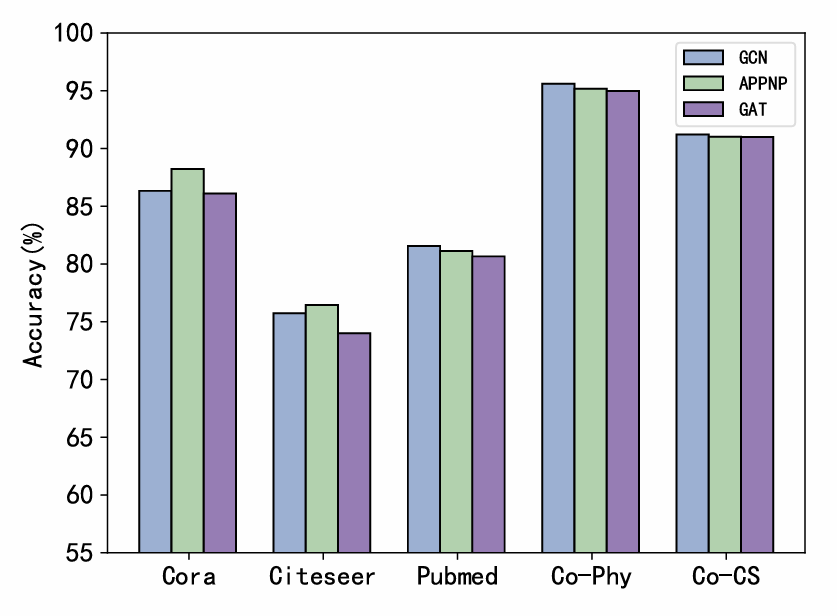}
  \caption{Comparison of SGBGC effects across different GNN models.}
  \label{fig:models}
\end{figure}

Figure \ref{fig:models} illustrates the robust performance of our SGBGC method across different GNN models. The SGBGC method effectively reduces the graph size while maintaining critical structural information and node features, resulting in high accuracy across various datasets. Notably, on complex datasets like Co-Phy and Co-CS, our method enabled different GNN models to achieve accuracies exceeding 93\%, demonstrating its effectiveness in preserving essential graph characteristics.

Additionally, our method achieved performance very close to or surpassing the best results on other datasets such as Pubmed and Cora, further validating its reliability and consistency. The ability of SGBGC to maintain high performance across different GNN architectures makes it a powerful tool for preprocessing graph data, ensuring scalability and efficiency in handling complex graph structures.

\subsection{Node Classification with Noisy Labels}
\label{fig:More Noise Experiments}

In experiments with different noise levels (5\%, 10\%, 15\%, and 20\%), SGBGC demonstrated higher robustness compared to other coarsening methods. The overall trends in Tables \ref{tab:classification_performance_5}, \ref{tab:classification_performance_10}, \ref{tab:classification_performance_15}, and \ref{tab:classification_performance_20} consistently show the superior performance of SGBGC.

Our noise experiment adopts uniform noise adding method mentioned in the paper\cite{55}, each node has the same probability to randomly use an wrong label, and this probability is the given noise rate. 

For instance, in the Cora dataset, SGBGC performed exceptionally well across all noise levels, achieving an accuracy of up to 85.57\% at \( r=0.5 \), significantly outperforming other methods. In the Citeseer dataset, SGBGC achieved an accuracy of over 72.41\% at \( r=0.1 \), at least 20\% higher than other methods. In the Pubmed dataset, SGBGC outperformed other methods across various \( r \) values, with the highest accuracy reaching 86.08\%. In the Co-CS and Co-Phy datasets, SGBGC showed outstanding performance across all noise levels and \( r \) values, with the highest accuracies reaching 93.52\% and 96.31\%, respectively.

The robustness of SGBGC is due to granular-ball computing, which mitigates label noise by clustering nodes based on purity thresholds. This preserves structural and feature integrity while reducing noise impact, resulting in high-quality representations and enhanced node classification accuracy. These results highlight the superior noise tolerance of SGBGC, making it a robust choice for real-world applications where data quality varies. 

\begin{table*}[tb!]
    \centering
    \caption{Comparison of SGBGC and other coarsening methods under different \(r\) values (5\% noise rate)}
    \label{tab:classification_performance_5}
    \begin{tabular}{lcccccc}
        \toprule
        DataSet & Ratio (\(r\)) & VNGC & VEGC & JCGC & GSGC & SGBGC \\
        \midrule
        \multirow{3}{*}{Cora} 
        & 0.5 & 77.03±1.50 & 79.66±1.19 & 79.64±1.91 & 81.43±1.23 & \textbf{81.96±1.28} \\
        & 0.3 & 77.15±1.66 & 76.04±1.84 & 75.99±1.94 & 80.15±1.59 & \textbf{81.82±1.33} \\
        & 0.1 & 67.84±1.52 & 58.66±3.29 & 63.84±1.68 & 64.35±1.78 & \textbf{82.34±0.74} \\
        \midrule
        \multirow{3}{*}{Citeseer} 
        & 0.5 & \textbf{73.20±1.42} & 72.11±1.12 & 70.13±2.75 & 72.19±1.30 & 72.13±0.93 \\
        & 0.3 & 71.07±1.77 & 68.96±1.76 & 68.42±2.51 & 62.11±4.10 & \textbf{75.20±0.86} \\
        & 0.1 & 62.18±3.19 & 47.18±1.34 & 52.70±3.98 & 55.89±4.72 & \textbf{72.46±0.83} \\
        \midrule
        \multirow{3}{*}{Pubmed} 
        & 0.3 & 83.99±0.19 & 83.22±0.27 & 81.95±0.63 & 83.45±0.21 & \textbf{86.94±0.22} \\
        & 0.1 & 81.69±0.42 & 82.79±0.22 & 82.19±0.56 & 80.20±2.70 & \textbf{84.84±0.29} \\
        & 0.05 & 64.69±0.86 & \textbf{82.64±0.39} & 81.91±0.84 & 76.49±2.98 & 82.20±0.43 \\
        \midrule
        \multirow{3}{*}{Co-CS} 
        & 0.3 & 56.20±5.16 & 53.37±5.93 & 48.40±4.26 & 51.03±5.04 & \textbf{93.18±0.23} \\
        & 0.1 & 38.38±3.13 & 33.63±0.07 & 33.16±1.65 & 43.74±3.11 & \textbf{91.73±0.30} \\
        & 0.05 & 23.79±0.64 & 33.37±0.65 & 23.77±2.78 & 33.24±0.36 & \textbf{91.05±0.28} \\
        \midrule
        \multirow{3}{*}{Co-Phy} 
        & 0.3 & 90.37±1.98 & 89.90±2.53 & 85.86±4.08 & 85.65±3.60 & \textbf{95.89±0.16} \\
        & 0.1 & 76.76±0.56 & 77.78±2.79 & 74.16±2.69 & 69.90±0.07 & \textbf{94.89±0.31} \\
        & 0.05 & 69.82±0.04 & 77.42±2.55 & 79.61±2.61 & 69.58±0.70 & \textbf{95.12±0.16} \\
        \bottomrule
    \end{tabular}
\end{table*}

\begin{table*}[tb!]
    \centering
    \caption{Comparison of SGBGC and other coarsening methods under different \(r\) values (10\% noise rate)}
    \label{tab:classification_performance_10}
    \begin{tabular}{lcccccc}
        \toprule
        DataSet & Ratio (\(r\)) & VNGC & VEGC & JCGC & GSGC & SGBGC \\
        \midrule
        \multirow{3}{*}{Cora} 
        & 0.5 & 77.54±1.41 & 78.99±1.45 & 79.22±1.52 & \textbf{80.96±2.00} & 77.36±1.43 \\
        & 0.3 & 77.01±1.38 & 76.03±2.19 & 75.09±1.38 & 77.77±2.94 & \textbf{78.66±1.51} \\
        & 0.1 & 70.00±1.32 & 59.67±2.86 & 64.24±2.62 & 63.74±2.14 & \textbf{81.41±1.13} \\
        \midrule
        \multirow{3}{*}{Citeseer} 
        & 0.5 & \textbf{73.81±0.08} & 70.57±1.54 & 69.44±2.98 & 69.34±1.30 & 68.94±1.11 \\
        & 0.3 & 70.62±2.31 & 68.05±2.69 & 66.41±3.68 & 66.41±3.68 & \textbf{72.88±0.86} \\
        & 0.1 & 63.47±4.52 & 41.63±2.32 & 52.95±4.90 & 43.39±5.57 & \textbf{72.22±0.93} \\
        \midrule
        \multirow{3}{*}{Pubmed} 
        & 0.3 & 83.95±0.23 & 82.53±0.46 & 82.38±1.04 & 82.95±0.28 & \textbf{86.77±0.29} \\
        & 0.1 & 80.84±0.67 & 82.72±0.44 & 82.30±0.41 & 82.37±0.77 & \textbf{84.61±0.33} \\
        & 0.05 & 67.94±5.58 & \textbf{83.07±0.55} & 82.47±0.48 & 79.04±2.77 & 82.36±0.55 \\
        \midrule
        \multirow{3}{*}{Co-CS} 
        & 0.3 & 55.59±5.95 & 50.87±5.20 & 45.32±3.32 & 51.10±3.85 & \textbf{92.58±0.23} \\
        & 0.1 & 34.55±1.91 & 33.72±0.11 & 33.11±2.50 & 40.30±3.85 & \textbf{91.92±0.26} \\
        & 0.05 & 24.09±0.93 & 33.72±0.04 & 33.08±2.14 & 30.45±5.00 & \textbf{91.08±0.30} \\
        \midrule
        \multirow{3}{*}{Co-Phy} 
        & 0.3 & 90.94±1.80 & 90.44±2.68 & 86.42±4.48 & 85.60±3.87 & \textbf{95.76±0.18} \\
        & 0.1 & 76.87±0.70 & 77.04±2.63 & 75.21±3.31 & 69.73±0.13 & \textbf{94.44±0.25} \\
        & 0.05 & 69.76±0.18 & 78.56±1.73 & 78.91±2.98 & 69.47±0.61 & \textbf{94.73±0.27} \\
        \bottomrule
    \end{tabular}
\end{table*}

\newpage

\begin{table*}[tb!]
    \centering
    \caption{Comparison of SGBGC and other coarsening methods under different \(r\) values (15\% noise rate)}
    \label{tab:classification_performance_15}
    \begin{tabular}{lcccccc}
        \toprule
        DataSet & Ratio (\(r\)) & VNGC & VEGC & JCGC & GSGC & SGBGC \\
        \midrule
        \multirow{3}{*}{Cora} 
        & 0.5 & 76.61±2.85 & 79.51±1.48 & 78.39±2.17 & \textbf{80.68±2.03} & 80.36±1.03 \\
        & 0.3 & 77.46±2.12 & 75.30±2.58 & \textbf{79.37±2.47} & 79.19±2.29 & 77.55±1.55 \\
        & 0.1 & 66.24±1.92 & 60.09±2.83 & 58.98±3.39 & 62.91±3.33 & \textbf{78.66±1.75} \\
        \midrule
        \multirow{3}{*}{Citeseer} 
        & 0.5 & \textbf{72.05±2.14} & 70.35±2.80 & 69.80±3.15 & 67.47±4.15 & 69.42±1.14 \\
        & 0.3 & \textbf{70.95±2.36} & 67.19±3.44 & 66.82±3.75 & 58.53±5.55 & 70.38±1.35 \\
        & 0.1 & 58.57±3.19 & 41.16±3.67 & 45.65±6.27 & 49.46±7.33 & \textbf{70.92±1.26} \\
        \midrule
        \multirow{3}{*}{Pubmed} 
        & 0.3 & 84.31±0.33 & 82.48±0.42 & 82.30±0.83 & 83.27±0.33 & \textbf{86.08±0.18} \\
        & 0.1 & 82.18±0.50 & 82.58±0.33 & 81.64±1.68 & 81.38±1.68 & \textbf{83.68±0.74} \\
        & 0.05 & 75.38±3.73 & 82.63±0.52 & 81.57±1.33 & 79.76±2.20 & \textbf{82.77±0.56} \\
        \midrule
        \multirow{3}{*}{Co-CS} 
        & 0.3 & 57.10±4.37 & 50.39±3.69 & 46.73±3.52 & 51.70±2.78 & \textbf{92.97±0.12} \\
        & 0.1 & 34.24±1.72 & 27.10±4.78 & 31.51±3.64 & 37.51±3.98 & \textbf{91.86±0.29} \\
        & 0.05 & 23.70±1.62 & 34.21±1.03 & 22.83±0.21 & 30.43±4.71 & \textbf{90.95±0.50} \\
        \midrule
        \multirow{3}{*}{Co-Phy} 
        & 0.3 & 91.05±1.91 & 90.16±2.93 & 85.69±3.34 & 84.32±3.73 & \textbf{96.15±0.05} \\
        & 0.1 & 75.64±0.89 & 78.90±3.44 & 77.42±3.65 & 69.53±0.31 & \textbf{95.64±0.13} \\
        & 0.05 & 75.08±0.75 & 79.77±1.42 & 82.44±3.33 & 65.39±7.19 & \textbf{95.11±0.21} \\
        \bottomrule
    \end{tabular}
\end{table*}

\begin{table*}[tb!]
    \centering
    \caption{Comparison of SGBGC and other coarsening methods under different \(r\) values (20\% noise rate)}
    \label{tab:classification_performance_20}
    \begin{tabular}{lcccccc}
        \toprule
        DataSet & Ratio (\(r\)) & VNGC & VEGC & JCGC & GSGC & SGBGC \\
        \midrule
        \multirow{3}{*}{Cora} 
        & 0.5 & 77.37±1.90 & 79.31±1.89 & 79.46±0.97 & \textbf{79.80±1.53} & 77.72±1.68 \\
        & 0.3 & 76.49±2.69 & 75.16±3.02 & 76.64±2.62 & \textbf{79.60±2.16} & 77.93±1.25 \\
        & 0.1 & 66.28±2.45 & 60.81±3.78 & 58.26±2.64 & 52.15±4.30 & \textbf{81.14±1.29} \\
        \midrule
        \multirow{3}{*}{Citeseer} 
        & 0.5 & \textbf{73.29±1.43} & 69.53±3.24 & 68.43±3.16 & 69.44±3.25 & 67.76±1.23 \\
        & 0.3 & 69.78±1.96 & 65.83±4.31 & 67.20±3.30 & 62.22±5.49 & \textbf{70.29±1.19} \\
        & 0.1 & 58.29±6.95 & 42.11±4.59 & 44.95±5.83 & 42.35±5.00 & \textbf{69.56±1.62} \\
        \midrule
        \multirow{3}{*}{Pubmed} 
        & 0.3 & 84.21±0.23 & 82.39±0.53 & 82.24±0.80 & 83.54±0.35 & \textbf{85.90±0.25} \\
        & 0.1 & 80.96±1.25 & 82.20±0.42 & 81.62±1.51 & 80.84±1.52 & \textbf{83.64±0.54} \\
        & 0.05 & 69.44±4.33 & 82.85±0.30 & 82.31±0.65 & 77.84±3.74 & \textbf{81.85±0.58} \\
        \midrule
        \multirow{3}{*}{Co-CS} 
        & 0.3 & 52.12±5.45 & 48.04±4.91 & 42.99±2.21 & 50.78±5.86 & \textbf{92.81±0.15} \\
        & 0.1 & 32.94±2.29 & 33.70±0.02 & 31.28±3.53 & 40.00±5.59 & \textbf{91.60±0.23} \\
        & 0.05 & 23.45±0.54 & 33.89±0.77 & 23.16±1.72 & 27.29±6.89 & \textbf{89.64±0.62} \\
        \midrule
        \multirow{3}{*}{Co-Phy} 
        & 0.3 & 91.24±1.65 & 91.18±2.14 & 86.74±3.07 & 85.31±4.95 & \textbf{95.98±0.12} \\
        & 0.1 & 76.47±0.49 & 76.80±2.04 & 77.92±4.35 & 69.87±1.29 & \textbf{95.49±0.14} \\
        & 0.05 & 72.03±2.98 & 75.34±3.16 & 81.09±3.01 & 62.19±8.41 & \textbf{94.91±0.21} \\
        \bottomrule
    \end{tabular}
\end{table*}


\clearpage
\onecolumn
\section{Algorithm}

\begin{algorithm}[!h]
\caption{Supervised Granular-Ball Graph Coarsening (SGBGC)}
\label{alg:SGBGC}
\begin{algorithmic}[1]
\Require $\mathcal{G} = (\mathcal{V}, \mathcal{E}, \mathbf{X})$; labels $\mathbf{Y}$
\Ensure $\tilde{\mathcal{G}} = (\tilde{\mathcal{V}}, \tilde{\mathcal{E}}, \tilde{\mathbf{X}})$; labels $\tilde{\mathbf{Y}}$
\State Initialize $\tilde{\mathcal{G}} = \emptyset$
\If{$\mathcal{G}$ is connected}
    \State \textcolor{gray}{\# For details on SGGBS, see Algorithm 3}
    \State $(\tilde{\mathcal{G}}, \tilde{\mathbf{Y}}) = \text{SGGBS}(\mathcal{G}, \mathbf{Y})$ 
\Else
    \State Decompose $\mathcal{G}$ into its connected components $\{\mathcal{G}_i\}$
    \For{each $\mathcal{G}_i$}
        \State $(\tilde{\mathcal{G}}_i, \tilde{\mathbf{Y}}_i) = \text{SGGBS}(\mathcal{G}_i, \mathbf{Y}|_{\mathcal{G}_i})$ 
    \EndFor
    \State Merge all $\tilde{\mathcal{G}}_i$ into $\tilde{\mathcal{G}}$ and concatenate corresponding $\tilde{\mathbf{Y}}_i$ into $\tilde{\mathbf{Y}}$
\EndIf
\State \Return $\tilde{\mathcal{G}}$, $\tilde{\mathbf{Y}}$
\end{algorithmic}
\end{algorithm}

\begin{algorithm}[!h]
\caption{Supervised Generation of Granular-Ball Subgraphs}
\label{alg:SGCGBG}
\begin{algorithmic}[1]
\Require Graph $\mathcal{G} = (\mathcal{V}, \mathcal{E}, \mathbf{X}, \mathbf{A})$, labels $\mathbf{Y}$, purity threshold $T$
\Ensure Set of granular-ball subgraphs $\mathcal{GB}_{s}$
\State Initialize the set of granular-ball subgraphs $\mathcal{GB}_{s} = \{\emptyset\}$
\State Define $\alpha = \sqrt{|\mathcal{V}|}$ where $|\mathcal{V}|$ is the number of nodes in $\mathcal{V}$
\State Select initial centers $\mathbf{c} = \{c_1, c_2, \ldots, c_{\alpha}\}$ based on highest degree nodes within each label category
\State Assign nodes to nearest center to form $\mathcal{GB}_{\text{init}} = \{\mathcal{GB}_1, \mathcal{GB}_2, \ldots, \mathcal{GB}_{\alpha}\}$
\For{$\mathcal{GB}_i \in \mathcal{GB}_{\text{init}}$}
    \State Set $\mathcal{GB}_{\text{current}} = \mathcal{GB}_i$
    \While{true}
        \State Find $c_1, c_2$ in $\mathcal{GB}_{\text{current}}$ that maximize the diversity of labels
        \State Assign non-central nodes to the nearest center to form two new granular-ball subgraphs $\mathcal{GB}_{c1}$ and $\mathcal{GB}_{c2}$ based on shortest path
        \If{$\text{purity}(\mathcal{GB}_{c1}) < T$ \textbf{or} $\text{purity}(\mathcal{GB}_{c2}) < T$}
            \State Split $\mathcal{GB}_{\text{current}}$ into $\mathcal{GB}_{c1}$ and $\mathcal{GB}_{c2}$
        \Else
            \State Add $\mathcal{GB}_{\text{current}}$ to $\mathcal{GB}_{s}$
            \State \textbf{break}
        \EndIf
    \EndWhile
\EndFor
\State \Return $\mathcal{GB}_{s}$
\end{algorithmic}
\end{algorithm}

\begin{algorithm}[!h]
\caption{Constructing the Granular-Ball Graph}
\label{alg:granular_ball_graph}
\begin{algorithmic}[1]
\Require $\mathcal{G} = (\mathcal{V}, \mathcal{E}, \mathbf{X})$, a list of granular-balls $\mathcal{GB}$
\Ensure Granular-ball graph $\tilde{\mathcal{G}}$

\State Initialize graph $\tilde{\mathcal{G}}$ with no nodes or edges
\State $\text{nodes} \leftarrow$ array of nodes in $\mathcal{G}$
\State $\text{node\_to\_GB\_index} \leftarrow$ empty map

\For{$i \leftarrow 1$ to $\text{length}(\mathcal{GB})$}
    \State Add node $i$ to $\tilde{\mathcal{G}}$ with label from $\mathcal{GB}[i]$
    \For{each $\text{node}$ in $\mathcal{GB}[i]$}
        \State $\text{node\_index} \leftarrow$ index of $\text{node}$ in $\text{nodes}$
        \State $\text{node\_to\_GB\_index}[\text{nodes}[\text{node\_index}]] \leftarrow i$
    \EndFor
\EndFor

\For{each edge $(u, v)$ in $\mathcal{E}$}
    \If{$u$ and $v$ in $\text{node\_to\_GB\_index}$}
        \State $u\_index \leftarrow \text{node\_to\_GB\_index}[u]$
        \State $v\_index \leftarrow \text{node\_to\_GB\_index}[v]$
        \If{$u\_index \neq v\_index$}
            \State Add edge $(u\_index, v\_index)$ to $\tilde{\mathcal{G}}$
        \EndIf
    \EndIf
\EndFor

\State \Return $\tilde{\mathcal{G}}$
\end{algorithmic}
\end{algorithm}

\end{document}